\documentclass{article}

\usepackage[preprint]{nips_2018}

\usepackage[utf8]{inputenc} 
\usepackage[T1]{fontenc}    
\usepackage{hyperref}       
\usepackage{url}            
\usepackage{booktabs}       
\usepackage{amsfonts}       
\usepackage{nicefrac}       
\usepackage{microtype}      
\usepackage{times}
\usepackage{latexsym}
\usepackage{subfig}
\usepackage[export]{adjustbox}
\usepackage{graphicx}
\usepackage{tabularx}
\usepackage{makecell}
\usepackage{amsmath}
\usepackage{float}
\usepackage{bm}
\usepackage{verbatim}
\usepackage{multirow}
\usepackage{color}
\usepackage[usenames, dvipsnames]{xcolor}

\usepackage{algorithm,algorithmicx}
\usepackage[noend]{algpseudocode}

\def\smallskip{\vspace{0mm}}

\newsavebox{\measurebox}

\title{Robust Neural Abstractive Summarization Systems and Evaluation against Adversarial Information}

\author{Lisa Fan$^{1}$ ~~~~ Dong Yu$^{2}$  ~~~~ Lu Wang$^{1}$\\
  $^{1}$College of Computer and Information Science, Northeastern University\\
  $^{2}$Tencent AI Lab\\
{\tt $^{1}$lisafan@ccs.neu.edu},~{\tt luwang@ccs.neu.edu}\\
{\tt $^{2}$dyu@tencent.com}
   \\}

\begin{document}

\maketitle

\date{}

\begin{abstract}
Sequence-to-sequence (seq2seq) neural models have been actively investigated for abstractive summarization. 
Nevertheless, existing neural abstractive systems frequently generate factually incorrect summaries and are vulnerable to adversarial information, suggesting a crucial lack of semantic understanding. 
In this paper, we propose a novel semantic-aware neural abstractive summarization model that learns to generate high quality summaries through semantic interpretation over salient content. 
A novel evaluation scheme with adversarial samples is introduced to measure how well a model identifies off-topic information, where our model yields significantly better performance than the popular pointer-generator summarizer. Human evaluation also confirms that our system summaries are uniformly more informative and faithful as well as less redundant than the seq2seq model. 

\end{abstract}

\section{Introduction}

Automatic text summarization holds the promise of alleviating the information overload problem~\citep{jones1999automatic}. Considerable progress has been made over decades, but existing summarization systems are still largely extractive---important sentences or phrases are identified from the original text for inclusion in the output~\citep{nenkova2011automatic}. Extractive summaries thus unavoidably suffer from redundancy and incoherence,
leading to the need for abstractive summarization methods. 
Built on the success of sequence-to-sequence (seq2seq) learning models~\citep{sutskever2014sequence}, there has been a growing interest in utilizing a neural framework for abstractive summarization~\citep{rush2015neural,nallapati2016abstractive,wang-ling:2016:N16-1,tan-wan-xiao:2017:Long,chen2018fast}. 
Although current state-of-the-art neural models naturally excel at generating grammatically correct sentences, the model structure and learning objectives have intrinsic difficulty in acquiring semantic interpretation of the input text, which is crucial for summarization. 
Importantly, the lack of semantic understanding causes existing systems to produce \textit{unfaithful generations}.
%
\citet{cao2017faithful} report that about 30\% of the summaries generated from a seq2seq model contain fabricated or nonsensical information. 

Furthermore, current neural summarization systems can be easily fooled by off-topic information. For instance, Figure~\ref{fig:adv_ex} shows one example where irrelevant sentences are added into an article about ``David Collenette's resignation''. 
Both the seq2seq attentional model~\citep{nallapati2016abstractive} and the popular pointer-generator model~\citep{see2017get} are particularly susceptible to unfaithful generation, partially because these models tend to rely on sentences at the beginning of the articles for summarization while being ignorant about their content. 
Therefore, we design a novel \textit{adversarial evaluation} metric to measure the robustness of each summarizer against small amounts of randomly inserted topic-irrelevant information. The intuition is that if a summarization system truly understands the salient entities and events, it would ignore unrelated content. 
To address the above issues, we propose a novel {\it semantic-aware abstractive summarization model}, inspired by the human process of writing summaries---important events and entities are first identified, and then used for summary construction. 
Concretely, taking an article as input, our model first generates a set of summary-worthy semantic structures consisting of predicates and corresponding arguments (as in semantic parsing), then constructs a fluent summary reflecting the semantic information.
Both tasks are learned under an encoder-decoder architecture with new learning objectives. 
A dual attention mechanism for summary decoding is designed to consider information from both the input article and the generated predicate-argument structures. 
We further present a novel decoder with a segment-based reranking strategy to produce diverse hypotheses and reduce redundancy under the guidance of generated semantic information. 

Evaluation against adversarial samples shows that while performance by the seq2seq attentional model and the pointer-generator model is impacted severely by even a small addition of topic-irrelevant information to the input, our model is significantly more robust and consistently produces more on-topic summaries (i.e. higher ROUGE and METEOR scores for standard automatic evaluation). 
Our model also achieves significantly better ROUGE and METEOR scores than both models on the benchmark dataset CNN/Daily Mail~\citep{hermann2015teaching}. 
Specifically, our model's summaries use substantially fewer and shorter extractive fragments than the comparisons and have less redundancy, alleviating another common problem for the seq2seq framework. 
Human evaluation demonstrates that our model generates more informative and faithful summaries than the seq2seq model.

\begin{figure}[t]
\centering
	\fontsize{9}{9}\selectfont
	\begin{tabularx}{\textwidth}{|X|}
    \hline
	\textbf{Article Snippet}: \textit{For years Joe DiMaggio was always introduced at Yankee Stadium as ``baseball's greatest living player.'' But with his memory joining those of Babe Ruth, Lou Gehrig, Mickey Mantle and Miller Huggins.}
    Canada's Minister of Defense resigned today, a day after an army official testified that top military officials had altered documents to cover up responsibility for the beating death of a Somali teen-ager at the hands of Canadian peacekeeping troops in 1993. Defense minister David Collenette insisted that his resignation had nothing to do with the Somalia scandal. \textit{Ted Williams was the first name to come to mind, and he's the greatest living hitter.} ...\\
\hline \hline
	
    \textbf{Seq2seq}: George Vecsey sports of The Times column on New York State's naming of late baseball legend Joe DiMaggio as ``baseball's greatest living player,'' but with his memory joining those of Babe Ruth, Lou Gehrig, Mickey Mantle and Miller dens.
 \\ 
     \textbf{Pointer-generator}: Joe DiMaggio is first name to come to mind, and Ted Williams is first name to come to mind, and he's greatest living hitter; he will be replaced by human resources minister, Doug Young, and will keep his Parliament seat for governing Liberal Party.

    \textbf{Our Model}: Former Canadian Defense Min David Collenette resigns day after army official testifies that top military officials altered documents to cover up responsibility for beating death of Somali teen-ager at hands of Canadian peacekeeping troops in 1993.
\\
	
	\hline
	\end{tabularx}
 	\vspace{-2mm}
	\caption{
    \fontsize{9}{10}\selectfont 
    Sample summaries for an adversarial example. 
    The inserted off-topic sentences are in \textit{italics}. The remainder of the input, truncated for space, consists entirely of the David Collenette story.
    }
\label{fig:adv_ex}
\end{figure}

\section{Related Work}
\label{sec:related}


To discourage the generation of fabricated content in neural abstractive models, a pointer-generator summarizer~\citep{see2017get} is proposed to directly reuse words from the input article via a copying mechanism~\citep{gu2016incorporating}. However, as reported in their work~\citep{see2017get} and confirmed by our experiments, this model produces nearly extractive summaries. 
While maintaining their model's rephrasing ability, here we improve the faithfulness and informativeness of neural summarization models by enforcing the generation of salient semantic structures via multi-task learning and a reranking-based decoder.

Our work is in part inspired by prior abstractive summarization work, where the summary generation process consists of a distinct content selection step (i.e., what to say) and a surface realization step (i.e., how to say it)~\citep{wang-cardie:2013:ACL2013,pighin-EtAl:2014:P14-1}. 
Our model learns to generate salient semantic roles and a summary in a single end-to-end trained neural network.

Our proposed model also leverages the recent successes of multi-task learning (MTL) as applied to neural networks for a wide array of natural language processing tasks~\citep{luong2015multi,sogaard2016deep,peng2017deep,rei:2017:Long,isonuma-EtAl:2017:EMNLP2017}. Most recent work~\citep{pasunuru2017towards} leverages MTL to jointly improve performance on summarization and entailment generation. Instead of treating the tasks equally, we employ semantic parsing for the sake of facilitating more informative and faithful summary generation.

\section{Model}
\label{sec:models}

\begin{figure*}
\centering
{\fontsize{10}{12}\selectfont
\sbox{\measurebox}{
\begin{minipage}[b]{.47\textwidth}
\centering
  \subfloat
    [Shared Decoder Model]
    {\label{fig:modelsA}\includegraphics[width=\textwidth]{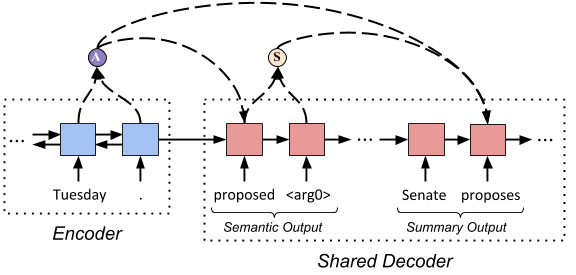}}
\vfill
\subfloat
  [Sample Input and System Outputs]
  {\label{fig:modelsB}\scriptsize 
  \setlength\tabcolsep{1.5pt}
  	\begin{tabular}{|ll|}
    \hline
    \textbf{Input:}&\\
    \textit{Article:}&{\fontfamily{phv}\selectfont the Senate proposed a tax bill on Tuesday .} \\
    \textbf{Output:}&\\
    \textit{Semantic:} &{\fontfamily{phv}\selectfont proposed {\tiny $<$ARG0$>$} the Senate {\tiny $<$ARG1$>$} tax bill} \\
    \textit{Summary:} &{\fontfamily{phv}\selectfont Senate proposes bill .}\\
    \hline
  	\end{tabular}}
\end{minipage}}
\usebox{\measurebox}\qquad
  \begin{minipage}[b][\ht\measurebox][s]{.47\textwidth}
\subfloat
  [Separate Decoder Model]
  {\label{fig:modelsC}\includegraphics[width=\textwidth]{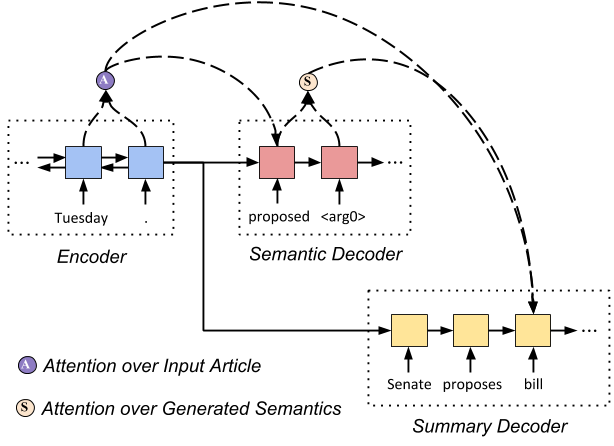}}
  \end{minipage}

\caption{
\fontsize{9}{10}\selectfont 
Our semantic-aware summarization model with dual attention over both input and generated salient semantic roles. 
In both shared and separate decoder models, the semantic roles are generated first, followed by the summary construction. Best viewed in color.}
\label{fig:models}}
\end{figure*}

\subsection{Model Formulation} \label{ssec:model_overview}
In the standard seq2seq model, a sequence of input tokens $\mathbf{x}=\{x_1,...,x_n\}$ is encoded by a recurrent neural network (RNN) with hidden states $\mathbf{h}_i$ at every timestep $i$. The final hidden state of the encoder is treated as the initial state of the decoder, another RNN ($\mathbf{h}_n=\mathbf{s}_0$). 
The decoder will produce the output sequence $\mathbf{y}=\{y_1,...,y_T\}$, with hidden states $\mathbf{s_t}$ at every timestep $t$. Our model outputs two sequences: the sequence of semantic role tokens $\mathbf{y}^s = \{y^s_1,...,y^s_{T^s}\}$ is generated first, followed by the sequence of summary (or abstract) tokens $\mathbf{y}^a = \{y^a_1,...,y^a_{T^a}\}$. 


The attention mechanism of \citet{bahdanau2014neural} is utilized to attend the input. Concretely, a context vector $\mathbf{c}^{\mathrm{inp}}_t$ is calculated as the summation of the encoder hidden states, weighted by the attention distribution $\mathbf{a}_t$ at every timestep $t$: 

\vspace{-4mm}
{\fontsize{10}{10}\selectfont
\setlength{\abovedisplayskip}{2pt}
\setlength{\belowdisplayskip}{2pt}
\begin{flalign}
\mathbf{c}^{\mathrm{inp}}_t =&\, \sum_{i=1}^{n}{a_{it}\mathbf{h}_i}  \label{eq:attn1}\\
\mathbf{a}_t =&\, \mathrm{softmax}(\mathbf{e}_t)  \label{eq:attn2}\\
e_{it} =&\, \mathbf{v}\tanh(\mathrm{W}_s\mathbf{s}_t+\mathrm{W}_h\mathbf{h}_i) \label{eq:attn3}
\end{flalign}} 
\noindent where $\mathbf{v}, \mathrm{W}_s, \mathrm{W}_h$ are learnable parameters.
The context vector, along with the decoder hidden state, is used to produce the vocabulary distribution via softmax:

\vspace{-3mm}
{\fontsize{10}{10}\selectfont
\setlength{\abovedisplayskip}{2pt}
\setlength{\belowdisplayskip}{2pt}
\begin{gather}
p(y_t\, | \,x_t,\mathbf{s}_t,\mathbf{c}^{inp}_t) = \mathrm{softmax}(\mathbf{o}_t) \\
\mathbf{o}_t = \mathrm{W}_v[\mathbf{s}_t\, \Vert \,\mathbf{c}^{\mathrm{inp}}_t]+\mathbf{d} \label{eq:output}
\end{gather}} 
\noindent where $\mathrm{W}_v$ and $\mathbf{d}$ are learned parameters and $\Vert$ represents vector concatenation.

The loss is defined as the negative log likelihood of generating summary $\mathbf{y}$ over the training set $D$ using model parameters $\theta$:

\vspace{-3mm}
{\fontsize{10}{10}\selectfont
\setlength{\abovedisplayskip}{2pt}
\setlength{\belowdisplayskip}{2pt}
\begin{gather} 
\mathrm{loss} = \sum_{(\mathbf{y}, \mathbf{x}) \in D}{-\log{p(\mathbf{y}\, | \,\mathbf{x};\theta)}}
\end{gather}}
The log probability for each training sample is the average log likelihood across decoder timesteps.

\smallskip
\noindent \textbf{Encoder.}
In our models, our encoder is a single-layer bidirectional long short-term memory (LSTM) unit \citep{graves2005framewise}, where the hidden state is a concatenation of the forwards and backwards LSTMs: $\mathbf{h}_i = [\overrightarrow{\mathbf{h}_i}\, \Vert \,\overleftarrow{\mathbf{h}_i}]$.

\smallskip
\noindent \textbf{Decoders.} 
We propose two different decoder architectures---\textit{separate decoder} and \textit{shared decoder}---to handle semantic information. In the separate decoder model (See Figure~\ref{fig:modelsC}), the \textit{semantic decoder} and the \textit{summary decoder} are each implemented as their own single-layer LSTM. 
This setup, inspired by the one-to-many multi-task learning framework of \citet{luong2015multi}, encourages each decoder to focus more on its respective task. 
Decoder output and attention over the input are calculated using Eqs.~\ref{eq:attn1}-\ref{eq:output}, but separate sets of parameters are learned for each decoder: $\mathbf{v}^s, \mathrm{W}^s_s, \mathrm{W}^s_h, \mathrm{W}_v^s, \mathbf{d}^s$ for the semantic decoder, and $\mathbf{v}^a, \mathrm{W}^a_s, \mathrm{W}^a_h, \mathrm{W}_v^a, \mathbf{d}^a$ for the summary decoder. 

We further study a shared decoder model (See Figure~\ref{fig:modelsA}) for the purpose of reducing the number of parameters as well as increasing the summary decoder's exposure to semantic information. One single-layer LSTM is employed to sequentially produce the important semantic structures, followed by the summary. 
Our output thus becomes $\mathbf{y} = [\mathbf{y}_s\, \Vert \,\mathbf{y}_a]$, and the first timestep of the summary decoder is the last timestep of the semantic decoder. 
Attention is calculated as in Eqs.~\ref{eq:attn1}-\ref{eq:output}. 


For both models, the loss becomes the weighted sum of the semantic loss and the summary loss: 

\vspace{-3mm}
{\fontsize{10}{10}\selectfont 
\setlength{\abovedisplayskip}{2pt}
\begin{flalign}\label{eq:loss}
\begin{split}
& \mathrm{loss} = -\sum_{(\mathbf{y}, \mathbf{x}) \in D} {\alpha\log{p(\mathbf{y}^a\, | \,\mathbf{x};\theta)}} 
	 +(1-\alpha)\log{p(\mathbf{y}^s\, | \,\mathbf{x};\theta)}
\end{split}
\end{flalign}} 
In our experiments, we set $\alpha$ as $0.5$ unless otherwise specified.

We also investigate {\it multi-head attention}~\citep{vaswani2017attention} over the input to acquire different language features. 
To our knowledge, we are the first to apply it to the task of summarization. As shown later in the results section, this method is indeed useful for summarization, with different heads learning different features. In fact, the multi-head attention is particularly well-suited for our shared decoder model, as some heads learn to attend semantic aspects and others learn to attend summary aspects.  

\smallskip
\noindent \textbf{Target Semantic Output.} 
We use semantic role labeling (SRL) as the target semantic representation, which identifies predicate-argument structures. To create the target data, articles in the training set are parsed in the style of PropBank \citep{kingsbury2002treebank} using the DeepSRL parser 
from \citet{he2017deep}.
We then choose up to five SRL structures that have the most overlap with the reference summary. Here we first consider matching headwords of predicates and arguments. If no match is found, we consider all content words. 
Note that the semantic labels are only used for model training. At test time, no external resource beyond the article itself is used for summary decoding.

\subsection{Dual Attention over Input and Semantics} \label{ssec:dual_attn}
To further leverage semantic information, a dual attention mechanism is used to attend over the generated semantic output in addition to the input article during summary decoding. 
Although attending over multiple sources of information has been studied for visual QA~\citep{nam2017dual} and sentence-level summarization~\citep{cao2017faithful}, these are computed over different {\it encoder states}. In contrast, our dual attention mechanism considers {\it decoding results} from the semantic output. 
Although our attention mechanism may appear close to intra- or self-attention~\citep{sankaran2016temporal}, the function of our dual attention is to attend over a specific portion of the previously generated content that represents its own body of information, whileas traditional self-attention is predominantly used to discourage redundancy.


The context vector attending over the semantic decoder hidden states is calculated as follows: 

\vspace{-3mm}
{\fontsize{10}{10}\selectfont
\setlength{\abovedisplayskip}{2pt}
\setlength{\belowdisplayskip}{2pt}
\begin{flalign}
\mathbf{c}^{\mathrm{sem}}_{t'} =&\, \sum_{j=1}^{T^s}{b_{jt'}\mathbf{s}^s_j}\\
\mathbf{b}_{t'} =&\, \mathrm{softmax}(\mathbf{f}_{t'})\\
f_{jt'} =&\, \mathbf{v}'\tanh(\mathrm{W}'_{s}\mathbf{s}^a_{t'}+\mathrm{W}'_{h}\mathbf{s}^s_j)
\end{flalign}} 
\noindent where $\mathbf{s}^a$ are the summary decoder's hidden states, $\mathbf{s}^s$ are the semantic decoder's hidden states, and $t'$ are the time steps for the summary decoder. The summary output $\mathbf{o}^a_{t'}$ now becomes:

\vspace{-3mm}
{\fontsize{10}{10}\selectfont
\setlength{\abovedisplayskip}{2pt}
\setlength{\belowdisplayskip}{2pt}
\begin{equation}
\mathbf{o}^a_{t'} = \mathrm{W}'_v[\mathbf{s}^a_{t'}\, \Vert \,\mathbf{c}^{\mathrm{inp}}_{t'}\, \Vert \,\mathbf{c}^{\mathrm{sem}}_{t'}]+\mathbf{d}' \label{eq:output2}
\end{equation}}

\subsection{Reranking-based Summary Decoder} \label{ssec:reranker} 
Albeit a standard practice, the beam search algorithm has been shown to produce suboptimal results in neural text generation problems~\citep{vijayakumar2016diverse} due to the fact that 
one beam often dominates others, yielding hypotheses that only differ by the last token. 
To combat this issue
, others have proposed beam rerankers that utilize external features such as Maximum Mutual Information \citep{li-EtAl:2016:N16-11}, segment-reranking \citep{shao-EtAl:2017:EMNLP2017}, and hamming loss \citep{wen-EtAl:2015:W15-46}. 
Here, we discuss our own segment-reranking beam search decoder, which encourages both {\it beam diversity} and {\it semantic coverage}. 
Our reranker is applied only during summary decoding, where we rely on the generated semantic roles for global guidance.
The only modification made to the semantic decoder is a de-duplication strategy, where generated predicates seen in previous output are eliminated. 

\smallskip
\noindent \textbf{Reranking.}
Regular beam search chooses the hypotheses for the next timestep solely based on conditional likelihood (i.e., $p(\mathbf{y}^a\, | \,\mathbf{x})$). 
In our proposed summary decoder, we also \textit{leverage our generated semantic information} and \textit{curb redundancy} during beam selection. Reranking is performed every $R$ timesteps based on the following scorer, where hypotheses with less repetition ($r$) and covering more content words from the generated semantics ($s$) are ranked higher:

\vspace{-3mm}
{\fontsize{10}{10}\selectfont
\setlength{\abovedisplayskip}{2pt}
\begin{flalign}
\mathrm{score} =&\, \log\left(p(\mathbf{y}^a\, | \,\mathbf{x})\right) + \alpha r + \beta s \\
r =&\, \log\left(1 - \frac{\textrm{\# LRS}*\vert\textrm{LRS}\vert}{\textrm{\# tokens in }\mathbf{y}^a}\right) 
\end{flalign}} 
%
where LRS is the longest repeating substring in the current hypotheses. 
We define a repeating substring as a sequence of three or more tokens that appears in that order more than once in the hypothesis, with the intuition that long repeating fragments (ex. ``{\fontfamily{lmtt}\selectfont the Senate proposed a tax bill; the Senate proposed a tax bill.}'') should be penalized more heavily than short ones (ex. ``{\fontfamily{lmtt}\selectfont the Senate proposed a tax bill; Senate proposed.}''). $s$ measures the percentage of generated semantic words reused by the current summary hypothesis, contingent on the predicate of semantic structure matching.
At every other timestep, we rank the hypotheses based on conditional likelihood and a weaker redundancy handler, $r'$, that considers unigram novelty (i.e., percentage of unique content words): $\mathrm{score} = \log\left(p(\mathbf{y}^a\, | \,\mathbf{x})\right) + \alpha' r'$. 
%
%
We use $R=10$, $\alpha=0.4$, $\beta=0.1$, $\alpha'=0.1$ in our experiments.

\smallskip
\noindent \textbf{Beam Diversity.}
In the standard beam search algorithm, all possible extensions to the beams are considered, resulting in a comparison of $B\times \mathcal{D}$ hypotheses, where $B$ is the number of beams and $\mathcal{D}$ is the vocabulary size. 
In order to make the best use of the reranking algorithm, we develop two methods to enforce hypothesis diversity during non-reranking timesteps. 

\smallskip
\noindent \textit{Beam Expansion.} 
Inspired by \citet{shao-EtAl:2017:EMNLP2017}, we rank only the $K$ highest scoring extensions from each beam, where $K<B$. This ensures that at least two unique hypotheses from the previous timestep will carry on to the next timestep. 

\smallskip
\noindent \textit{Beam Selection.} We further reinforce hypothesis diversity through the two-step method of (1) likelihood selection and (2) dissimilarity selection. In likelihood selection, we accept $N$ hypotheses (where $N<B$) from our pool of $B\times K$ based solely on conditional probabilities, as in traditional beam search. From the remaining hypotheses pool, we select $B-N$ hypotheses on the basis of dissimilarity. We choose the hypotheses with the highest dissimilarity score $\Delta([h]_N, h')$, where $h'$ is a candidate and $[h]_N$ are the hypotheses chosen during likelihood selection. In our experiments, we use token-level Levenshtein edit distance \citep{levenshtein1966binary} as the dissimilarity metric, where ${\Delta([h]_N, h') = \max_{h \in [h]_N}\left(\mathrm{Lev}(h, h')\right)}$. 
In experiments, we use $B=12$, $K=6$, $N=6$.

\section{Experimental Setup}
\label{sec:exp}

\noindent \textbf{Datasets.}
We experiment with two popular large datasets of news articles paired with human-written summaries: the CNN/Daily Mail corpus~\citep{hermann2015teaching} (henceforth CNN/DM) and the New York Times corpus~\citep{sandhaus2008new} (henceforth NYT). 
For CNN/DM, we follow the experimental setup from \citet{see2017get} and obtain a dataset consisting of 287,226 training pairs, 13,368 validation pairs, and 11,490 test pairs.
%
For NYT, we removed samples 
with articles of less than 100 words or summaries of less than 20 words. We further remove samples with summaries containing information outside the article, e.g., ``[AUTHOR]'s movie review on...'' where the author's name does not appear in the article. 
NYT consists of 280,146 training pairs and 15,564 pairs each for validation and test. 





\smallskip
\noindent \textbf{Training Details and Parameters.} 
For all experiments, a vocabulary of 50k words shared by input and output is used. 
%
Model parameters and learning rate are adopted from prior work~\citep{see2017get} for comparison purpose. 
All models are also trained in stages of increasing maximum token lengths to expedite training. 
The models trained on the NYT dataset use an additional final training stage, where we optimized only on the summary loss (i.e., $\alpha=1$ in Eq.~\ref{eq:loss}).
During decoding, unknown tokens are replaced with the highest scoring word in the corresponding attention distribution. 

%

\smallskip
\noindent \textbf{Baselines and Comparisons.}
We include as our extractive baselines \textsc{Textrank} \citep{mihalcea2004textrank} and \textsc{lead-2}, the first 2 sentences of the input article, simulating the average length of the target summaries. 
We consider as abstractive comparison models (1) vanilla seq2seq with attention (\textsc{seq2seq}) and (2) pointer-generator~\citep{see2017get} (\textsc{Point-Gen}), which is trained from scratch using the released code. 
Results for variants of our model are also reported.\footnote{Reinforcement learning methods have been recently proposed for abstractive summarization~\citep{paulus2018deep,celikyilmaz2018deep,chen2018fast}. We will study this type of learning models in future work.}

\smallskip
\noindent \textbf{Automatic Evaluation Metrics.} 
For automatic evaluation, we first report the F1 scores of ROUGE-1, 2, and L~\citep{Lin:2003:AES:1073445.1073465}, and METEOR scores based on exact matching and full matching that considers paraphrases, synonyms, and stemming~\citep{denkowski2014meteor}. 

We further measure two important aspects of summary quality: \emph{extractiveness}---how much the summary reuses article content verbatim, and \emph{redundancy}---how much the summary repeats itself. 
For the first aspect, we utilize the density metric proposed by \citet{grusky2018newsroom} that calculates ``the average length of extractive fragments'':

\vspace{-3mm}
{\fontsize{10}{10}\selectfont
\setlength{\abovedisplayskip}{1pt}
\begin{flalign}
\textsc{Density}(A,S) =&\, \frac{1}{\vert S\vert}\sum_{f \in F(A,S)}\vert f \vert^2 \label{eq:density}
\end{flalign}}
where $A$ represents the article, $S$ represents the summary, and $F(A,S)$ is a set of greedily matched extractive fragments from the article-summary pair. 

Based on density, we propose a new redundancy metric:

\vspace{-3mm}
{\fontsize{10}{10}\selectfont
\setlength{\abovedisplayskip}{1pt}
\begin{flalign}
\textsc{Redundancy}(S) =&\, \frac{1}{\vert S\vert}\sum_{f' \in F^\prime (S)}(\#f'\times \vert f^\prime \vert)^2 \label{eq:redundancy}
\end{flalign}}
where $F'(S)$ contains a set of fragments at least three tokens long that are repeated within the summary, and $\#f^\prime$ is the repetition frequency for fragment $f^\prime$. Intuitively, longer fragments and more frequent repetition should be penalized more heavily. 

\section{Results}
\label{sec:results}


\smallskip
\noindent \textbf{Adversarial Evaluation.} 
Our pilot study suggests that the presence of minor irrelevant details in a summary often hurts a reader's understanding severely, but such an error cannot be captured or penalized by the recall-based ROUGE and METEOR metrics. 

In order to test a model's ability to discern irrelevant information, we design a novel adversarial evaluation scheme, where we purposely mix a limited number of off-topic sentences into a test article. The intuition is that if a summarization system truly understands the salient entities and events, it would ignore unrelated sentences. 

We randomly select 5,000 articles from the CNN/DM test set with the ``news'' tag in the URL (mainly covering international or domestic events).
For each article, we randomly insert one to four sentences from articles in the test set with the ``sports'' tag. 
For NYT, we randomly select 5,000 articles from the test set with government related tags (``U.S.'', ``Washington'', ``world'') and insert one to four sentences from articles outside the domain (``arts'', ``sports'', ``technology''). We ensure that sentences containing a pronoun in the first five words are not interrupted from the previous sentence so that discourse chains are unlikely to be broken. 
the content of the article, it would ignore unrelated sentences. We took 15,000 articles from the NYT corpus with government related tags (``u.s.'', ``washington'', ``world''), and for each article, randomly inserted up to four sentences from articles outside the domain (``arts'', ``sports'', ``technology''). We ensured sentences that contained a pronoun in the first five words were not interrupted from the previous sentence so as not to break discourse or coreference chains.

\begin{figure}[ht]
\centering
\fontsize{5}{12}\selectfont
\subfloat[CNN/DM ROUGE-L]{\label{fig:adv_meteor}\includegraphics[width=.25\columnwidth,valign=b]{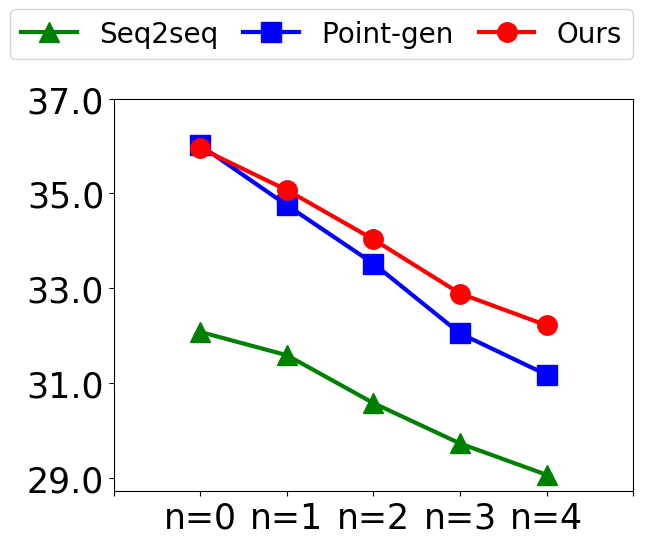}}
\subfloat[CNN/DM METEOR]{\label{fig:adv_meteor}\includegraphics[width=.25\columnwidth,valign=b]{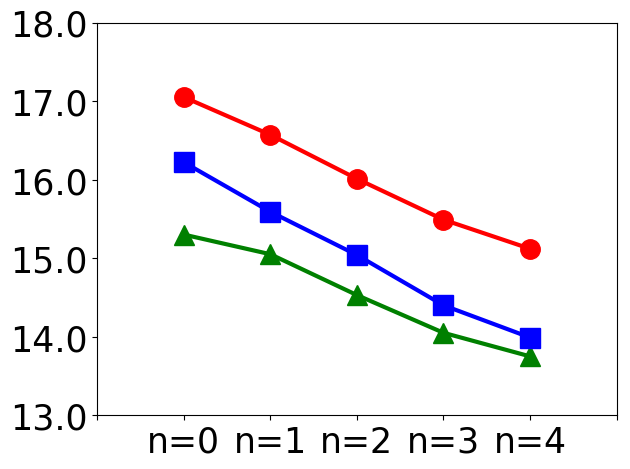}}
\subfloat[NYT ROUGE-L]{\label{fig:adv_rouge}\includegraphics[width=.25\columnwidth,valign=b]{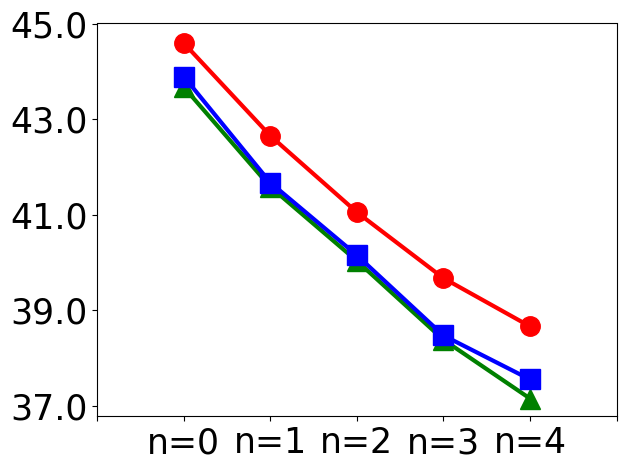}}
\subfloat[NYT METEOR]{\label{fig:adv_meteor}\includegraphics[width=.25\columnwidth,valign=b]{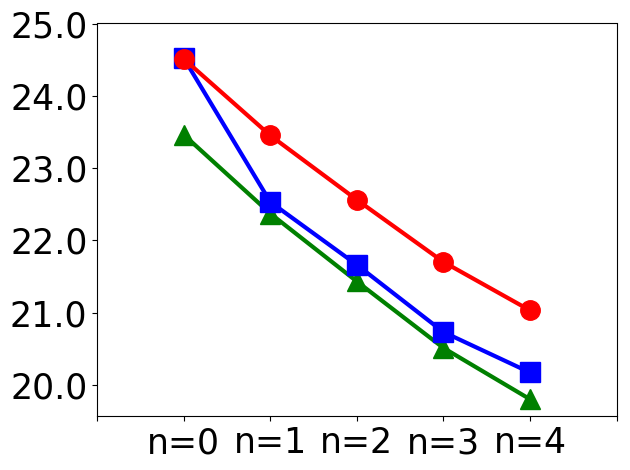}}
\caption{\fontsize{9}{10}\selectfont 
ROUGE-L and full METEOR scores on adversarial samples, where $n$ irrelevant sentences are inserted into original test articles. Our models (shared decoder for NYT, shared+MHA for CNN/DM) are sturdier against irrelevant information than seq2seq and pointer-generator.  
}
\label{fig:adv_eval}
\end{figure}

The adversarial evaluation results on seq2seq, pointer-generator, and our shared decoder model are shown in Figure~\ref{fig:adv_eval}, where our model consistently yields significantly better ROUGE-L and METEOR scores than the comparisons, and is less affected as more irrelevant sentences are added. Sample summaries for an adversarial sample are shown in Figure~\ref{fig:adv_ex}. 
We find that our semantic decoder plays an important role in capturing salient predicates and arguments, leading to summaries of better quality. 

\begin{table*}[t]
\centering
{
\fontsize{9}{10}\selectfont
\setlength{\tabcolsep}{0.7mm}
\begin{tabular}{l||r r r|r r|r r||r r r|r r|r r}
\hline & \multicolumn{7}{c||}{CNN/DailyMail}& \multicolumn{7}{c}{NYT}\\


\hline & \multicolumn{3}{c|}{\bf ROUGE} & \multicolumn{2}{c|}{\bf METEOR} & 
\multicolumn{1}{c}{\bf Dens.} &\multicolumn{1}{c||}{\bf Red.}& \multicolumn{3}{c|}{\bf ROUGE} & \multicolumn{2}{c|}{\bf METEOR} & 
\multicolumn{1}{c}{\bf Dens.} &\multicolumn{1}{c}{\bf Red.} \\ \cline{2-6}\cline{9-13} 
& \multicolumn{1}{c}{R-1} & \multicolumn{1}{c}{R-2} & \multicolumn{1}{c|}{R-L} & \multicolumn{1}{c}{exact} & \multicolumn{1}{c|}{full}& \multicolumn{2}{c||}{} & \multicolumn{1}{c}{R-1} & \multicolumn{1}{c}{R-2} & \multicolumn{1}{c|}{R-L} & \multicolumn{1}{c}{exact} & \multicolumn{1}{c|}{full}& \multicolumn{2}{c}{} \\ \hline

\textsc{Human}  & \multicolumn{1}{c}{-}& \multicolumn{1}{c}{-}& \multicolumn{1}{c|}{-}& \multicolumn{1}{c}{-}& \multicolumn{1}{c|}{-} 
& 3.40 & 0.06& \multicolumn{1}{c}{-}& \multicolumn{1}{c}{-}& \multicolumn{1}{c|}{-}& \multicolumn{1}{c}{-}& \multicolumn{1}{c|}{-}
& 4.32 & 0.12\\
\textsc{Lead-2}  & 38.35 & 15.74 & 31.48 & 14.44&15.62
& 59.19&  0.33 & 39.01 & 21.96 & 30.84 &  16.73 & 17.91 
& 68.31 & 0.12\\
\textsc{Textrank} & 33.78 & 12.41 & 28.75 & 12.73 & 13.94 
& 36.44& 7.94 & 34.81 & 16.02 & 25.56 & 14.69 & 16.14   
& 49.11& 0.65\\
\hline\hline
\multicolumn{15}{l}{\it Abstractive Comparisons}\\
\textsc{Seq2Seq}   &  32.14 & 12.33 & 29.41 & 13.66 & 14.65
&8.46 & 11.92 & 44.53 & 28.54 & 37.05 & 19.62 & 20.54 
& 7.31 & 3.09 \\ 
\textsc{Point-Gen}   & 35.60 & 15.24 & 32.43 & 15.50 & 16.54 & 
16.80 & 10.77 & 46.76 & 31.13 & 38.85 & 20.62 & 21.55 & 
9.54 & 2.67 \\
\hline\hline
\multicolumn{15}{l}{\it Our Semantic-Aware Models}\\
\textsc{Dec$_{share}$} & 35.44 &  14.18 &  32.28 &  14.80  & 15.84 & 
12.42 & 5.97& \textbf{46.40} &  \textbf{30.55} &  \textbf{38.47} &  \textbf{19.91}  & \textbf{20.79}
& 8.73 & \textbf{0.83}\\
\quad + \textsc{MHA} &  \textbf{36.31} &  14.96 &  \textbf{33.21} &  \textbf{15.56}  & \textbf{16.62} 
& 13.03 & \textbf{5.20} &45.88 &  30.26 &  38.23  & 19.32 &  20.17
& \textbf{8.33}  &  1.29  \\ 
\textsc{Dec$_{sep}$} & 35.31 &  13.97 &  32.23 &  14.81 &  15.86 & 
\textbf{11.55} &5.21   &45.98 &  30.23  & 38.27 &  19.47  & 20.33 & 
8.68 & 1.10 \\
\quad + \textsc{MHA} & 36.07 & \textbf{15.09} &  33.08 &  15.25&   16.29 & 
12.93 & 5.65 &46.10 &  30.37  & 38.41 &  19.43  & 20.29 
&8.58 & 0.89\\  \hline
\end{tabular}
\caption{\fontsize{9}{10}\selectfont  
Results on CNN/Daily Mail and NYT. For our model, we display four variants, based on shared or separate decoder, and with or without multi-head attention (MHA). In addition to ROUGE and METEOR, 
we also display extractive density (Dens.) and redundancy (Red.) (lower scores are preferred). Best performing amongst our models are in {\bf bold}. 
All our ROUGE scores have a 95\% confidence interval of at most $\pm$0.04. 
Our models all statistically significantly outperform seq2seq for ROUGE and METEOR (approximate randomization test, $p<0.01$). Our shared decoder with MHA model statistically significantly outperforms pointer-generator in ROUGE-1 and ROUGE-L on CNN/DM.
}
\label{tab:results}
}
\end{table*}

\noindent \textbf{Automatic Evaluation.} 
The main results are displayed in Table~\ref{tab:results}. 
%
On CNN/DM, all of our models significantly outperform seq2seq across all metrics, and our shared decoder model with multi-head attention yields significantly better ROUGE (R-1 and R-L) scores than the pointer-generator on the same dataset (approximate randomization test, $p<0.01$). 
Despite the fact that both ROUGE and METEOR favor recall and thus reward longer summaries, our summaries that are often shorter than comparisons still produce significantly better ROUGE and METEOR scores than seq2seq on the NYT dataset. 


Furthermore, our system summaries, by all model variations, \emph{reuse fewer and shorter phrases from the input} (i.e., lower density scores) than the pointer-generator model, signifying a potentially stronger ability to rephrase. Note that the density scores for seq2seq are likely deflated due to its inability to handle out-of-vocabulary words.
Our models also produce \emph{less redundant} summaries than their abstractive comparisons. 
\smallskip
\noindent \textbf{Human Evaluation.}
We further conducted a pilot human evaluation on 60 samples selected from the NYT test set. For each article, the summaries by the seq2seq model, our shared decoder model, and the human reference were displayed in a randomized order. 
Two human judges, who are native or fluent English speakers, were asked to read the article and rank the summaries against each other based on non-redundancy, fluency, faithfulness to input, and informativeness (whether the summary delivers the main points of the article). Ties between system outputs were allowed but discouraged. 

As illustrated in Table~\ref{tab:humaneval}, our model was ranked significantly higher than seq2seq across all metrics. 
Surprisingly, our model's output was ranked higher than the reference summary in 26\% of the samples for informativeness and fluency. We believe this is due to the specific style of the NYT reference summaries and the shorter lengths of our summaries: 
the informal style of the reference summaries (e.g., frequently dropping subjects and articles) negatively affects its fluency rating, and readers find our shorter summaries to be more concise and to the point. 


\begin{table}[ht]
{
\centering
\fontsize{9}{10}\selectfont
\setlength{\tabcolsep}{0.8mm}
\begin{tabular}{l c c c c}
\hline &\multicolumn{1}{c}{\textsc{Non-Red.}} &\multicolumn{1}{c}{\textsc{Fluency}}&\multicolumn{1}{c}{\textsc{Faith.}}& \multicolumn{1}{c}{\textsc{Inform.}} \\ \hline
\textsc{Human} & 1.4 $\pm$ 0.7& 1.5 $\pm$ 0.7& 1.5 $\pm$ 0.8  & 1.6 $\pm$ 0.8 \\
\textsc{Seq2Seq} & 2.0 $\pm$ 0.8& 2.4 $\pm$ 0.7& 2.1 $\pm$ 0.8 &  2.4 $\pm$ 0.7 \\
\textsc{Ours}& 1.6 $\pm$ 0.7& 2.1 $\pm$ 0.8& 1.9 $\pm$ 0.7  &  2.0 $\pm$ 0.7  \\ \hline
\end{tabular}
\vspace{1mm}
\caption{\fontsize{9}{10}\selectfont  
Human ranking results on non-redundancy, fluency, faithfulness of summaries, and informativeness. The mean ($\pm$ std. dev.) for the rankings is shown (lower is better). Across all metrics, the difference between our model (shared decoder) and human summary, as well as between our model and seq2seq is statistically significant (one-way ANOVA, $p<0.05$). 
}
\label{tab:humaneval}
}
\end{table}

\vspace{-2mm}
\section{Discussion}
\label{sec:discussion}
\noindent \textbf{Usage of Semantic Roles in Summaries.} 
We examine the utility of the generated semantic roles.
Across all models, approximately 44\% of the generated predicates are part of the reference summary, indicating the adequacy of our semantic decoder. 
Furthermore, across all models, approximately 65\% of the generated predicates are reused by the generated summary, and approximately 53\% of the SRL structures are reused by the system using a strict matching constraint, in which the predicate and head words for all arguments must match in the summary. 
When gold-standard semantic roles are used for dual attention in place of our system generations, ROUGE scores increase by about half a point, indicating that improving semantic decoder in future work will further enhance the summaries.

\smallskip
\noindent \textbf{Coverage.} 
We also conduct experiments using a coverage mechanism similar to the one used in \citet{see2017get}. We apply our coverage in two places: (1) over the input to handle redundancy, and (2) over the generated semantics to promote its reuse in the summary. 
However, no significant difference is observed. Our proposed reranker handles both issues in a more explicit way, and does not require the additional training time used to learn coverage parameters. 

\smallskip
\noindent \textbf{Alternative Semantic Representation.} 
Our summarization model can be trained with other types of semantic information. For example, 
in addition to using the salient semantic roles from the input article, 
we also explore using SRL parses of the reference abstracts as training signals, but the higher level of abstraction required for semantic generation hurts performance by two ROUGE points for almost all models, indicating the type of semantic structure matters greatly for the ultimate summarization task. 

For future work, other semantic representation along with novel model architecture will be explored.
For instance, other forms of semantic representation can be considered, such as frame semantics \citep{baker1998berkeley} or Abstract Meaning Representation (AMR) \citep{banarescu2013abstract}. Although previous work by \citet{vinyals2015grammar} has shown that seq2seq models are able to successfully generate linearized tree structures, we may also consider generating semantic roles with a hierarchical semantic decoder \citep{sordoni2015hierarchical}.

\section{Conclusion}
\label{sec:conclusion}
We presented a novel semantic-aware neural abstractive summarization model that jointly learns summarization and semantic parsing. A novel dual attention mechanism was designed to better capture the semantic information for summarization. A reranking-based decoder was proposed to promote the content coverage. 
Our proposed adversarial evaluation demonstrated that our model was more adept at handling irrelevant information compared to popular neural summarization models. 
Experiments on two large-scale news corpora showed that our model yielded significantly more informative, less redundant, and less extractive summaries. 
Human evaluation further confirmed that our summaries were more informative and faithful than comparisons. 





\newpage
{
\bibliography{references}
\bibliographystyle{acl_natbib_nourl}
}

\newpage
\setcounter{section}{0}
\renewcommand{\thesection}{\Alph{section}}
\section*{Supplementary Materials}
\section{NYT preprocessing}
We base our preprocessing method on those taken by \citet{paulus2017deep}.
We first removed samples without a summary, as well as articles with less than 100 words and corresponding summaries of less than 20 words. The Stanford CoreNLP Tokenizer \citep{manning-EtAl:2014:P14-5} is used to parse the articles and summaries. All words are made lowercase except those that are named entities (parsed with Stanford CoreNLP), and replace all numbers with ``0''. Template words at the ends of summaries are removed: ``(s)'', ``(m)'', ``photo'', ``graph'', ``chart'', ``map'', ``table'', and ``drawing''. We then removed samples where the summary contained the following in the first 5 words: ``article'', ``column'', ``op-ed'', ``essay'', ``editorial'', ``letter'', ``profile'', ``interview'', ``excerpts'', ``news'', ``analysis'', ``review''. These words were indicative of a template-style summary containing information outside articles, e.g., ``[AUTHOR]'s movie review of...'' where the author's name does not appear in the article. While this trimmed about 30\% of our data, we found that our method may have been too conservative, as some column names longer than 5 words (ex. ``Lisa Belkin life 's work column on ...'') were not successfully omitted.

This gives us a dataset of 311,274 samples, where we put the samples in chronological publication order and took 280,146 samples (90\%) for the training set, and 15,564 samples (5\%) each for validation and test sets.

\section{Target Semantic Output Creation}
We developed two algorithms for creating the target semantic outputs: strict matching and soft matching. Soft matching was only used for samples where the strict matching produced no results. In both cases, we first parse the article in the style of PropBank \citep{kingsbury2002treebank} using \citeauthor{he2017deep}'s (\citeyear{he2017deep}) DeepSRL parser, and keep only the predicates, ARG0 (the agent), ARG1 (the patient), and ARG2 (instrument, benefactive, or attribute).

For the NYT dataset, 80\% of the samples produced target SRL structures using strict matching, while only 0.01\% of samples did not have any target SRL structures after the soft matching operation. Each sample had approximately 5 SRL structures per article.

\paragraph{Strict Matching.}
We first perform stemming on the output words using Stanford CoreNLP and consider only the last word in each argument. We define the head word to be the last word in the argument. We then perform the following matching algorithm for each SRL structure parsed from the article:
\begin{algorithm}
\begin{algorithmic}
\State \textit{pred} = predicate
\State \textit{Stop} = [list of stop words]
\State \textit{Summ} = [list of summary words]
\If {\textit{pred} $\in$ \textit{Stop} \textbf{or} \textit{pred} $\not\in$ \textit{Summ}}
	\State \textsc{Reject}
\Else 
	\If {$\exists$ ARG0 \textbf{and} ARG0 $\not\in$ \textit{Stop}}
    	\If {ARG0 $\in$ \textit{Summ}}
        	\State \textsc{Accept}
        \Else
        	\State \textsc{Reject}
        \EndIf
    \ElsIf {$\exists$ ARG1 \textbf{and} ARG1 $\not\in$ \textit{Stop}}
    	\If {ARG1 $\in$ \textit{Summ}}
        	\State \textsc{Accept}
        \Else
        	\State \textsc{Reject}
        \EndIf
    \Else
        \State \textsc{Reject}
    \EndIf
\EndIf
\end{algorithmic} 
\end{algorithm}

\paragraph{Soft Matching.}
We first perform stemming on the output words using Stanford CoreNLP. We then count the number of unique non-stop words in the predicate, ARG0, and ARG1 that are also in the summary, and choose at most the top 5 SRL structures with at least two matched words in order of number of matched words.

\section{Multi-head Attention over Input} \label{ssec:multihead_attn}
Multi-head attention~\citep{vaswani2017attention} over the input has been used in various NLP tasks to acquire different language features. 
Concretely, $K$ separate context vectors, or heads, attend over the input, and the final context vector $\mathbf{c}^{inp}_t$ takes the form of a linear transformation over the concatenation of all heads.
\noindent Each context vector $\mathbf{c}_t^i$ is calculated using Eqs.~\ref{eq:attn1}-\ref{eq:attn3}, with independent parameters $\mathbf{v}^i$ and $\mathrm{W}^i_h$. All attention heads share the same parameter $\mathrm{W}_s$ that transforms the decoder hidden state $\mathbf{s}_t$.
\begin{flalign}
&\mathbf{c}_t^i = \sum_{j}^{j=n}{a^i_{jt}\mathbf{h}_j}\\
&\mathbf{a}^i_t = \mathrm{softmax}(\mathbf{e}^i_t)\\
&e^i_{jt} = \mathbf{v}^i\tanh(\mathrm{W}_s\mathbf{s}_t+\mathrm{W}^i_h\mathbf{h}_j)
\end{flalign}

\section{Redundancy Handling}

In our proposed beam search summary decoder, reranking is performed every $R$ timesteps based on the following scorer, where hypotheses with less repetition ($r$) and covering more content words from the generated semantics ($s$) are ranked higher:

{\fontsize{10}{10}\selectfont
\setlength{\abovedisplayskip}{2pt}
\begin{flalign}
\mathrm{score} =&\, \log\left(p(\mathbf{y}^a\, | \,\mathbf{x})\right) + \alpha r + \beta s \\
r =&\, \log\left(1 - \frac{\textrm{\# LRS}*\vert\textrm{LRS}\vert}{\textrm{\# tokens in }\mathbf{y}^a}\right) \\
s =&\, \frac{\textrm{\# unique argument tokens in }\mathbf{y}^s\textrm{ and }\mathbf{y}^a}{\textrm{\# unique tokens in }\mathbf{y}^s}
\end{flalign}}
where LRS is the longest repeating substring in the current hypotheses. We define a repeating substring as a sequence of three or more tokens that appear 
more than once in the hypothesis, with the intuition that long extractive fragments should be penalized more heavily than short ones. Our longest repeating substring algorithm is implemented by finding the deepest non-leaf node in a suffix tree. 
At all other timesteps, we rank the hypotheses based on conditional likelihood and a weaker redundancy handler:

{\fontsize{10}{10}\selectfont
\setlength{\abovedisplayskip}{2pt}
\begin{flalign}
\mathrm{score} =&\, \log\left(p(\mathbf{y}^a\, | \,\mathbf{x})\right) + \alpha' r' \\
r' =&\, \frac{\textrm{\# unique tokens in }\mathbf{y}^a}{\textrm{\# tokens in }\mathbf{y}^a}
\end{flalign}}
We omit stop words for $r'$ and $s$.

\section{Finding Set of Repeating Fragments}
We describe the procedure for finding the set of self-repeating substrings used in the redundancy automatic evaluation metric.\\
1. Extract all repeated substrings using a suffix tree, and only keep substrings $\geq$ 3 tokens.\\
2. Sort by number of occurrences, most to least.\\
3. Within the same number of occurrences, sort by substring length, longest to shortest.\\
4. For each substring in the sorted substrings, if the substring has no overlap with any items in results, add the substring to the results. Here, overlap is defined as ``x is a substring of y'' or ``y is a substring of x''.

The above procedure favors number of occurrences over length of repeated substring.

~

\begin{figure}[t]
\centering
	\fontsize{8.5}{9}\selectfont
    \small
	\begin{tabularx}{\columnwidth}{|X|}
    \hline
	\textbf{Article Snippet}: Carmen Villegas did not expect the Roman Catholic Archdiocese of New York to send burly guards into her church, Our Lady Queen of Angels, but it did. She did not expect to be at the center of a chaotic scene, shouting at church officials and flinging open the front door of the church, but she was. She did not expect to be arrested, but she was, after she refused to leave. And she did not expect the archdiocese to close her beloved East Harlem church, two weeks early, but late Monday night, that is exactly what it did. Ms. Villegas and some fellow parishioners occupied the sanctuary for about 28 hours , protesting the archdiocese's plan to shut down Our Lady Queen of Angels on March 1. ...\\
    \textbf{Human Abstract}: Roman Catholic Archdiocese of New York closes Our Lady Queen of Angels church in East Harlem two weeks earlier than planned after protests and vigil by parishioners lead to arrests. \\
    \hline \hline
    \textbf{Seq2seq}: \textit{Carmen [UNK], Roman Catholic Archdiocese of New York}, and five other pastor, Carmen [UNK], are evicted from church's East Harlem church after refusing to leave. \\ 
    \textbf{Our Model}: Catholic \textcolor{ForestGreen}{Archdiocese} of New York is \textcolor{ForestGreen}{closing} its \textcolor{ForestGreen}{East Harlem church}, two weeks early; \textcolor{violet}{Villegas and some fellow parishioners} \textcolor{blue}{occupy sanctuary} for about 28 hours, \textcolor{red}{protesting archdiocese's plan to shut down Our Lady Queen of Angels on March 1}. \\
    
    \quad \textbf{Accompanied SRL Output}: \\
    \quad {\tiny $<$PRED$>$} \textcolor{ForestGreen}{close} {\tiny $<$ARG0$>$} \textcolor{ForestGreen}{the archdiocese} {\tiny $<$ARG1$>$} her \textcolor{ForestGreen}{East}\\
    \qquad \textcolor{ForestGreen}{Harlem church} \\
    \quad {\tiny $<$PRED$>$} \textcolor{blue}{occupied} {\tiny $<$ARG0$>$} \textcolor{violet}{Ms. Villegas and some fellow} \\
    \qquad \textcolor{violet}{parishioners} {\tiny $<$ARG1$>$} \textcolor{blue}{the sanctuary} \\
    \quad {\tiny $<$PRED$>$} \textcolor{red}{protesting} {\tiny $<$ARG0$>$} \textcolor{violet}{Ms. Villegas and some}  \\
    \qquad \textcolor{violet}{fellow parishioners} {\tiny $<$ARG1$>$} \textcolor{red}{the archdiocese's plan} \\
    \qquad \textcolor{red}{to shut down Our Lady Queen of Angels on March 1} \\

	\hline
	\end{tabularx}

	\caption{\fontsize{10}{12}\selectfont
    Sample summaries by our model and seq2seq. Errors are in \textit{italics}. Our model generates salient semantic roles, which guide the construction of informative and correct summary. Semantic reusage by our summary is shown in color.
    }
\label{fig:example}
\end{figure}


\section{Example Output}
Figure~\ref{fig:example} illustrates an example of our system summary. Our semantic decoder is able to generate structures that reuse arguments.

\section{Training Details}
\paragraph{Experimental Setup.}
For all experiments, a vocabulary of 50k words shared by input and output is used. Word embeddings of size 128 are randomly initialized and learned during training. 
The single-head attention vector and hidden states of LSTMs used in all models have 256 dimensions. 
When multi-head attention is employed, four attention heads are learned, each of size 64. The Adagrad optimizer~\cite{duchi2011adaptive} is used with initial accumulator 0.1 and a learning rate of 0.15. We use a batch size of 16. The training process took 3 to 5 days depending on model and GPU type (both Quadro P5000 GPU and Tesla V100 GPU were used). Models all converge after 10 to 19 epochs. 

\paragraph{Training in Stages.}
To expedite training time, we train all models in stages of increasing maximum token lengths---starting from highly truncated maximum lengths, we then increase the lengths when the models converge. Specifically, the three stages are of input/output token lengths 50/50, 200/50, and 400/100, where semantic and summary outputs each are truncated to the output token length. The models trained on the NYT dataset used a fourth training stage, in which we optimized only on the summary loss (ie. $\alpha=1$ in Eqn 8).

\paragraph{Dealing with Unknown Tokens.}
In our models, we deal with the generated unknown tokens by replacing them with the highest scoring word in the corresponding attention distribution, since the generated unknown tokens very often correspond to out-of-vocabulary proper nouns. In multi-head attention models, we choose the highest scoring word in the summed attention distributions. This replacement method was shown to boost ROUGE scores by at least 1 point across all models.

\paragraph{Initializing with Seq2seq.}
We only include target SRL structures that appear in the system input. Therefore, for the highly truncated training stage, many training samples do not include a target semantic output. To ensure the model saw all the data, all semantic-aware models were initialized with the weights from the converged first stage baseline seq2seq model when possible ($\mathbf{v}^s, \mathrm{W}^s_s, \mathrm{W}^s_h, \mathrm{W}_v^s, \mathbf{d}^s$ are still initialized randomly for the separate decoder models). Both our original and modified decoders required the generated summary and semantic output to be at least 35 tokens long.

\section{Generated SRL Statistics}

\begin{table}[t]
\centering
{
\fontsize{9}{10}\selectfont
\setlength{\tabcolsep}{0.6mm}
\begin{tabular}{|l|l|l|l|l|}
\hline && \multicolumn{3}{c|}{\% generated / avg length}  \\ \cline{3-5} 
 & \multicolumn{1}{c|}{\# SRL}  & \multicolumn{1}{c|}{arg0} & \multicolumn{1}{c|}{arg1} & \multicolumn{1}{c|}{arg2}\\ \hline
reference & 3.8 & 62.2 / 4.0 & 91.6 / 6.5 & 20.2 / 6.5\\
shared\_dec       & 3.4 & 66.6 / 5.0 & 95.3 / 8.8 & 20.1 / 8.9 \\ \hline
\end{tabular}
\vspace{1mm}
\caption{\fontsize{10}{12}\selectfont Statistics on generated SRL.}
\label{tab:gen_srl}
}
\end{table}

We found that our semantic decoder generates fewer SRL structures than the average target SRL parse, but generates a similar ratio of the individual arguments (See Table~\ref{tab:gen_srl}). The average length of our generated arguments is also longer than in the target SRL structures. In fact, the decoder will often first generate a structure with very long arguments, then generate subsequent structures from portions of the previous argument. For example, after generating ARG1 in the last SRL structure of Figure~\ref{fig:example}, the semantic decoder would likely next generate a structure with the predicate ``shut'' (as in ``shut down''). This is partially a byproduct of imperfect target semantic parsings that exhibit a similar behavior. Improving the trade-off between argument lengths and the number of semantic structures (ie. generating more succinct semantic structures) is one direction to explore in future work.

\end{document}